\documentclass[sigconf]{acmart}
\AtBeginDocument{%
  }

\copyrightyear{2026}
\acmYear{2026}
\setcopyright{cc}
\setcctype{by}
\acmConference[AI Summit '26]{ACM AI Summit 2026}{August 31-September 02, 2026}{Atlanta, GA, USA}
\acmBooktitle{ACM AI Summit 2026 (AI Summit '26), August 31-September 02, 2026, Atlanta, GA, USA}
\acmDOI{10.1145/3806096.3844817}
\acmISBN{979-8-4007-2638-5/2026/08}

\acmISBN{978-1-4503-XXXX-X/2018/06}

\usepackage{tasks}

\newcommand\syllogism[3][]{%
  \begin{center}
  \def\tmp{#1}%
  \ifx\tmp\empty\else(#1)\quad\fi
  \begin{tabular}{@{}l@{}}
    #2\\\hline#3\quad$\therefore$
  \end{tabular}
  \end{center}
}
\begin{document}

\title{Non-Zero Radius: The Missing Component of Neural Networks 
}

\author{Tiansi Dong$^{1,2}$, Mateja Jamnik$^{1}$, Pietro Liò$^{1}$} 
\affiliation{%
  \institution{$^{1}$Computer Lab, The University of Cambridge\\ $^{2}$The Alan Turing Institute }
  \city{Cambridge \& London}
 \country{United Kingdom}
}
\email{\{td540,mj201,pl219\}@cam.ac.uk, tdong@turing.ac.uk}

\renewcommand{\shortauthors}{Dong et al.}

\begin{abstract} 
Inspired by the analogy between the {\em all-or-none} activation of biological neurons and the true-false of logic circuits, McCulloch and Pitts proposed a mathematical model of the biological neuron in 1943 with the aim of simulating logical reasoning. Although Norbert Wiener considered the model overly simplistic at the time, it later became a building block of modern neural networks, including large language models. Despite their remarkable empirical success, however, conventional neural networks remain largely black-box systems and struggle with reliable symbolic reasoning.
Viewing vectors as spheres of zero radius, we relax the {\em zero-radius} constraint and represent concepts as spheres with non-zero radii. We argue that sphere-based representations can capture additional features of biological cognition, including {\em representing spatial concepts as regions} and {\em reasoning through model construction and inspection}. Building on sphere embeddings, we introduce Sphere Neural Networks (SphNNs), which use a transition map to achieve symbolic-level Aristotelian syllogistic reasoning without training data, thereby taking an initial step toward the original goal of McCulloch and Pitts. We suggest that the zero-radius constraint may reflect an aspect of the oversimplification identified by Wiener and may limit conventional neural networks’ capacity for symbolic-level logical reasoning.
\end{abstract}


\begin{teaserfigure}
 \includegraphics[width=\textwidth]{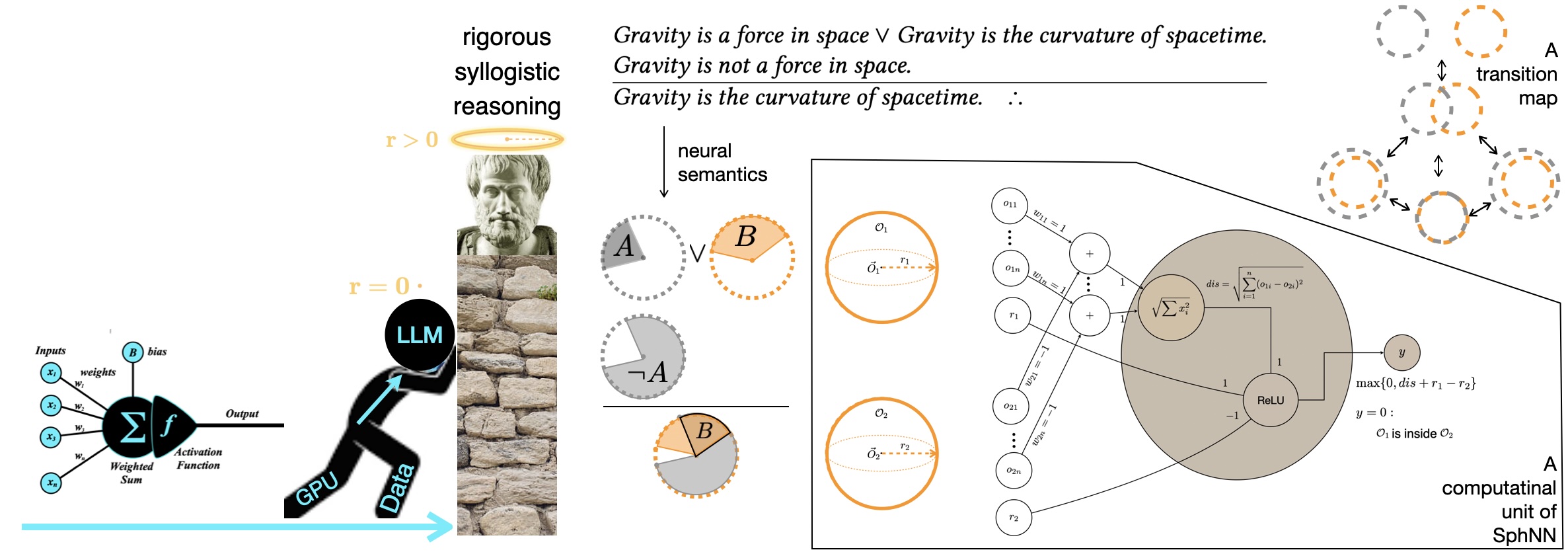}
  \caption{Large language models (LLMs) still struggle with reliable logical reasoning. Conventional neural networks represent concepts as vectors, which can be viewed as spheres of zero radius. Sphere Neural Networks (SphNNs) represent concepts as spherical regions with non-zero radii. These regions and their geometric transformations, guided by a transition map, enable symbolic-level Aristotelian syllogistic reasoning without training data.}
  \label{fig:teaser}
\end{teaserfigure}

\received{30 June 2026}
\received[revised]{ }
\received[accepted]{ }

\maketitle

\section{Introduction}
On February 19, 2026, Demis Hassabis, Co-Founder and CEO of Google DeepMind, suggested the following test of Artificial General Intelligence (AGI): 
``One way we maybe would test that is you could imagine training a foundation model with a knowledge cutoff of something like 1911 and then see if it could come up with general relativity like Einstein did in 1915.'' Passing this test would require, among other abilities, basic logical inference such as the following:
\syllogism{{\em Gravity is a force in space} $\lor$ {\em Gravity is the curvature of spacetime.}\\ {\em Gravity is not a force in space.}}{{\em Gravity is the curvature of spacetime.}}
Reasoning of the form ``A$\lor$ B. $\neg$ A. $\therefore$ B'' is known as a disjunctive syllogism. As a foundational form of logical deduction, Aristotelian syllogistic reasoning dominated the study of logic for more than two millennia, and prominent psychologists have regarded it as a microcosm of human rational intelligence~\citep{Khemlani2021}. 
It is ubiquitous in everyday life and underpins reasoning across a wide range of real-world domains, including education~\citep{education15}, business~\citep{syladvertising14}, medical decision-making~\citep{syllobio2025}, and legal judgement~\citep{LJA23,law23}.

Observing that biological neurons exhibit an “all-or-none” firing behaviour analogous to the operation of logical circuits, Warren S. McCulloch and Walter Pitts proposed the first mathematical model of neural information processing in 1943~\cite{McCulloch1943}. The paper has only three references: Principia Mathematica~\cite{russell25}, The Logical Syntax of Language~\cite{carnap1937}, and Principles of Theoretical Logic~\cite{hilbert38}. McCulloch and Pitts linked neural activity to formal logic. Syllogistic reasoning provides a basic test of this connection.

At that time, Warren S. McCulloch was an established neuroscientist and psychiatrist, while Walter Pitts was a remarkably gifted young autodidact. After publishing their seminal paper, both became associated with MIT. Norbert Wiener regarded their neural model as overly simplistic, but he also recognised Pitts’s extraordinary talent and invited him to participate in his doctoral seminar, providing an opportunity to further develop the mathematical model. Unfortunately, tensions that arose during the collaboration between McCulloch and Wiener eventually strained Pitts's relationship with Wiener as well. Pitts withdrew increasingly from academic life and struggled with alcoholism in his later years. He died at age 45, leaving behind a profound legacy as a pioneer of computational neuroscience and artificial intelligence~\cite{pittslife}. 
What, exactly, was oversimplified? As illustrated in Figure~\ref{fig:teaser}, extending point representations to spheres enables symbolic-level syllogistic reasoning. Could the missing element in the original model be the radius?

\section{The McCulloch–Pitts' Model Falls Short of  Symbolic-Level Syllogistic Reasoning}

\begin{figure}[!t] 
\centering \includegraphics[width=0.5\linewidth]{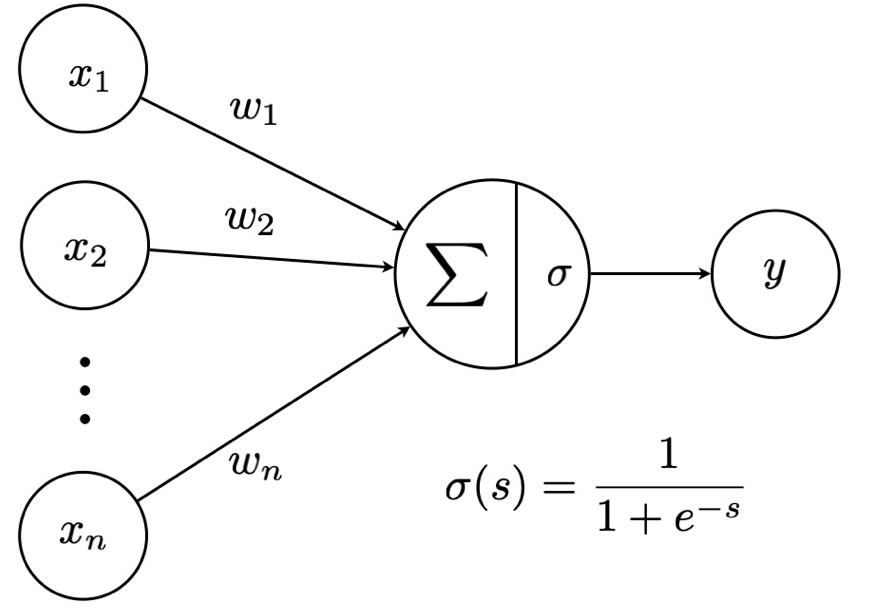}
\caption{The input of a traditional perceptron is a vector.}
\label{fsph1}
\end{figure}

The simple mathematical neuron proposed by McCulloch and Pitts laid the foundation for connectionist approaches to artificial intelligence. Building upon this framework, Rosenblatt introduced the perceptron~\citep{rosenblatt1958} (Figure~\ref{fsph1}), 
which subsequently became the fundamental computational unit of modern neural networks. Over the following decades, perceptron-based architectures evolved into increasingly sophisticated models, including multilayer neural networks~\citep{Rumelhart86xor}, Convolutional Neural Networks (CNNs)\citep{LeNet,ImageNet_Nips12}, Recurrent Neural Networks (RNNs)\citep{Hochreiter19rudder,LeCunNature15}, Transformer architectures~\citep{Vaswani17}, and, more recently, Large Language Models (LLMs)~\citep{palm22023,llama22023,openAI23,mistral7b,mistral23}. Despite their impressive performance across diverse tasks (as in, e.g.,~\citep{alphaGo2020,AlphaGeo,AlphaFold3}), conventional neural networks still face fundamental challenges in logical reasoning, make errors in simple abstract reasoning~\citep{Eisape2024,syllogism24}, and provide correct answers with incorrect explanations~\citep{creswell2022selectioninference,zelikman2022star}. It remains unclear whether LLMs truly reason~\citep{chatgpt_nature2023,Mitchell_science.adj5957}. 

Several studies have explored the syllogistic reasoning performance of LLMs~\cite{Eisape2024,syllogism24,syllobio2025}. 
Among the evaluated PaLM~2~\citep{palm22023} and Llama~2~\citep{llama22023} models, \mbox{PaLM~2-Small} outperformed \mbox{PaLM~2-Large}, showing that syllogistic reasoning accuracy did not increase consistently with model size~\cite{Eisape2024}.
Additionally, LLMs in the PaLM 2 family and GPT-3.5~\citep{openAI23} may achieve above-chance performance in familiar situations but exhibit numerous imperfections in abstract reasoning, including syllogisms~\cite {syllogism24}; Mistral LLMs~\citep{mistral7b,mistral23}, Gemma LLMs~\citep{gemma2024}, Llama-3 LLMs~\citep{llama3}, and BioMistral LLMs~\citep{biomistral2024} achieved average zero-shot accuracies of 70\% on {\em generalised modus ponens} and
23\% on {\em disjunctive syllogism}, and both zero- and few-shot LLMs are sensitive to surface-level lexical variations~\cite{syllobio2025}. The evaluated LLMs fall short of the reliability required for high-stakes biomedical applications, let alone attaining the rigour of symbolic-level reasoning. 
These findings highlight persistent limitations in conventional neural networks’ logical reasoning~\cite{djl2026b}. 
 
\section{Mental Model Construction with Regions}

Beyond the ``all-or-none’’ principle, spatial cognition offers another source of inspiration: Tolman’s pioneering behavioural experiments demonstrated that rats form cognitive maps of their environment~\cite{Tolman48}. In these cognitive maps, spatial objects are represented as regions, while spatial hierarchies are naturally captured by nested regions, allowing explicit representation of parent–child relationships~\citep{Davis81,McDermott81,StevensCoupe78,McNamara86}. Subsequent advances in neuroscience identified a zoo of spatial cells in the hippocampus, including grid cells, place cells, and boundary cells~\citep{keefe78,Savelli08,Solstad08,Lever09}. More recently, evidence suggests that mental spatial representations also encode positions in abstract dimensions of experience, constructing cognitive spaces for domain-general reasoning. Geometric accounts of these cognitive spaces represent properties and concepts as convex regions~\citep{Bellmundeaat6766,tversky19}.

Empirical evidence strongly supports the view that rational reasoning proceeds through model construction and inspection~\cite{LairdByrne91,knauf03,GoodwinLaird05,Knauff09,Knauff13}: {\em 
What we have is a procedure to represent and integrate in our mind the content of premises by means of models comparable to schematic pictures of the situation. We then read the conclusions of these models}~\cite{enigma2018}. 
In mental model theory, reasoning proceeds through {\em model construction, model inspection, and model variation}  \cite{LairdByrne91}. In the {\em model variation} phase, people try to construct alternative models to refute the conclusion. 
If a counter-model is found (the premises are true and the conclusion is false), the inference is {\em invalid}. The preferred mental model theory argues that people construct a preferred and simplified model in mind, in a deterministic manner, initially setting aside other possibilities \cite{RagniMarkus13,Knauff13} -- In this account, initial model construction is deterministic~\cite{RagniMarkus13}. Subsequent models are generated by local transformations following the principle of minimal changes~\cite{Harman1986,Gardenfors1988,GAeRDENFORS90,knauffetc13}, with transitions represented by a neighbourhood graph~\cite{GoodayCohn94,DongJPL}.

\section{Vectors are Zero-Radius Spheres}
One extension of McCulloch-Pitts' model was the {\em diameter-limited perceptron} whose input signals $\vec{x}_i$ are restricted within a certain fixed diameter \cite[p.12]{minsky88}: Each $\vec{x}_1,\dots,\vec{x}_n$ is located within a sphere.

\begin{figure}[!t] 
\centering \includegraphics[width=1\linewidth]{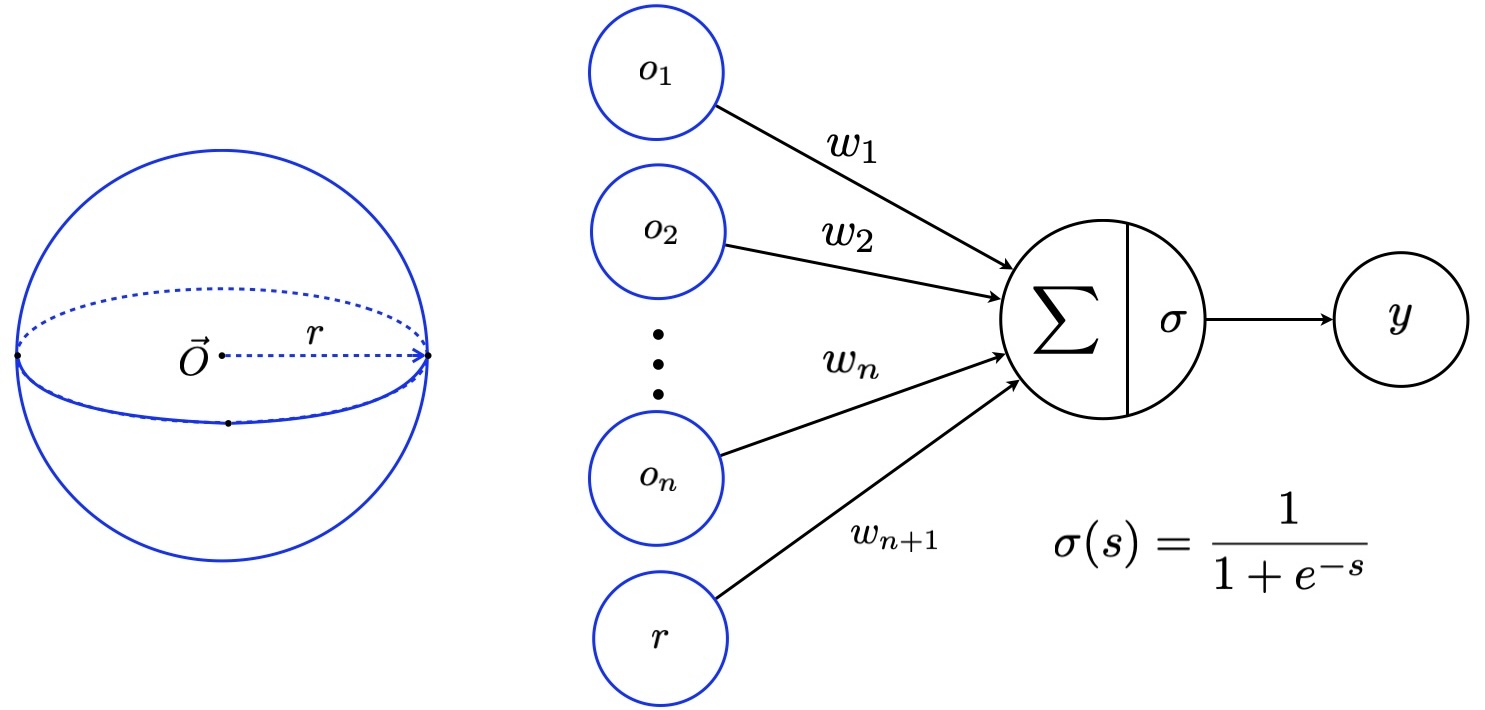}
\caption{The input of a diameter-limited perceptron is restricted inside a sphere with the centre $\vec{O}$ and the radius $r$.}
\label{fsphere1}
\end{figure} 

We further elaborate on this idea by explicitly defining this sphere, whose existence is consistent with convex regions in mental representation~\cite{Bellmundeaat6766}, and also with McCulloch-Pitts' model: vectors are special spheres with zero radius. This extends conventional neural networks to Sphere Neural Networks (SphNNs)~\cite{djl2024sphere,djl2025}. The input of a SphNN is the input domain of a {\em diameter-limited perceptron} $[o_1, \dots, o_n, r]$, where $[o_1, \dots, o_n]$ is the central vector $\vec{O}$ and $r \ge 0$ is the radius, such that $\|\vec{x}_i-\vec{O}\|\le r$ ($1\le i \le n$). 

\begin{figure}[!t]
\centering \includegraphics[width=1\linewidth]{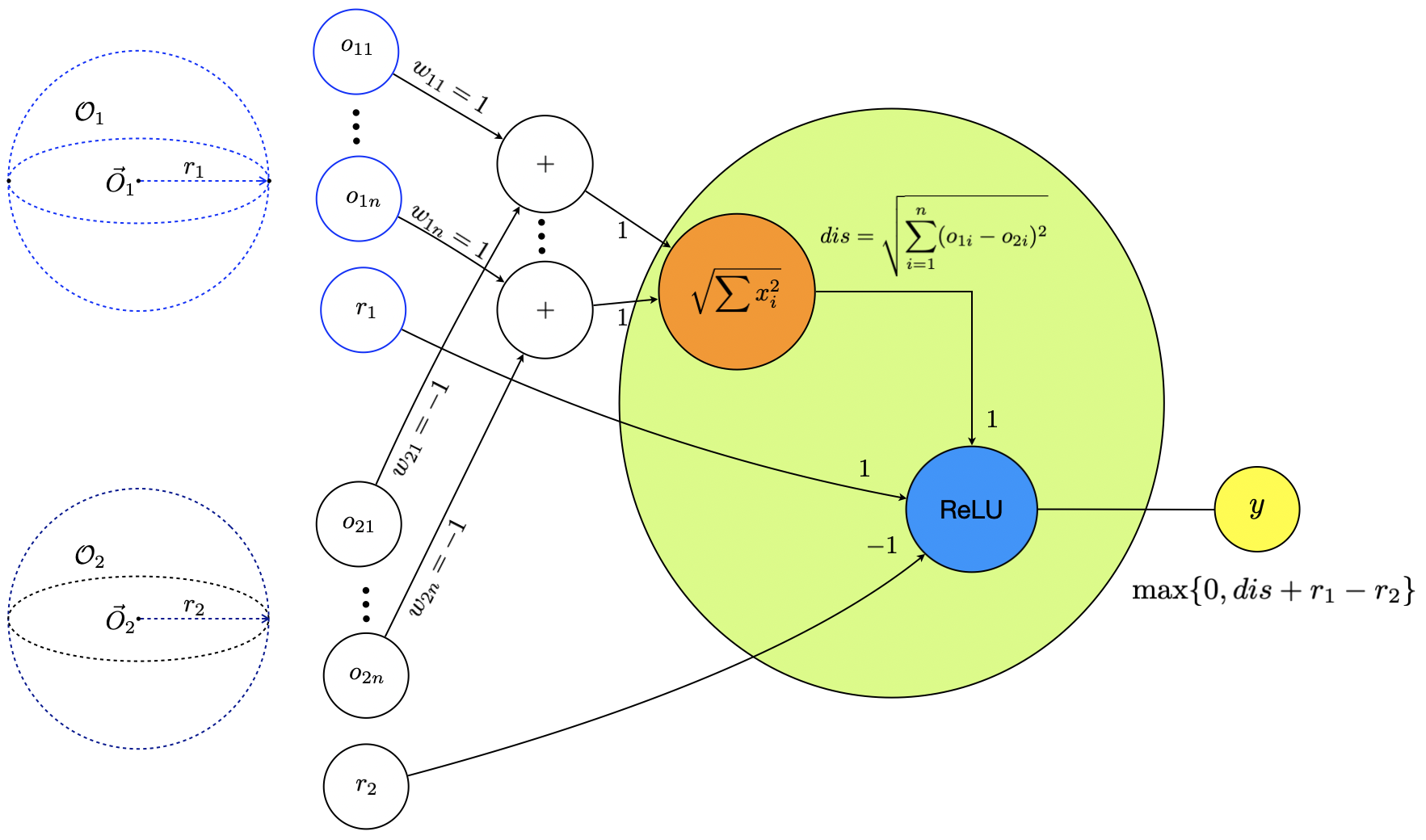} \caption{A non-monotonic activation function signals whether $\mathcal{O}_1$ is inside $\mathcal{O}_2$.}
\label{fsphere2}
\end{figure}

Biological neurons can be distinguished through their output encodings~\cite{Ballard86}: one type of neuron outputs {\em frequency} signals, simulated by {\em perceptron} \cite{minsky88,Anderson95}, the other type of neurons outputs {\em spatial} or {\em place} encoding, for example, neurons in the visual cortex \cite{Ballard86}, whose output indicates the degree of matching between the input stimuli and the receptive field. The non-zero-radius extension 
provides a geometric model of receptive-field matching.
We may represent the input stimuli and the receptive field as two spheres $\mathcal{O}_1$ and $\mathcal{O}_2$, with the non-monotonic activation function $\Delta(\mathcal{O}_1,\mathcal{O}_2)\triangleq \max\{0, \sqrt{\sum_{i=1}^n(o_{1i}-o_{2i})^2} + r_1 - r_2\}$, as illustrated in Figure~\ref{fsphere2}. When $\Delta(\mathcal{O}_1,\mathcal{O}_2)=0$, $\mathcal{O}_1$ is contained in $\mathcal{O}_2$; when $r_1 = r_2$, the larger $\Delta(\mathcal{O}_1,\mathcal{O}_2)$ is, the farther apart the centres of $\mathcal{O}_1$ and  $\mathcal{O}_2$ are.\looseness-1

\section{Representing Set-Theoretic Relations and Syllogistic Statements as Sphere Relations}

\begin{figure}[!t]
\centering 
\includegraphics[width=0.48\textwidth]{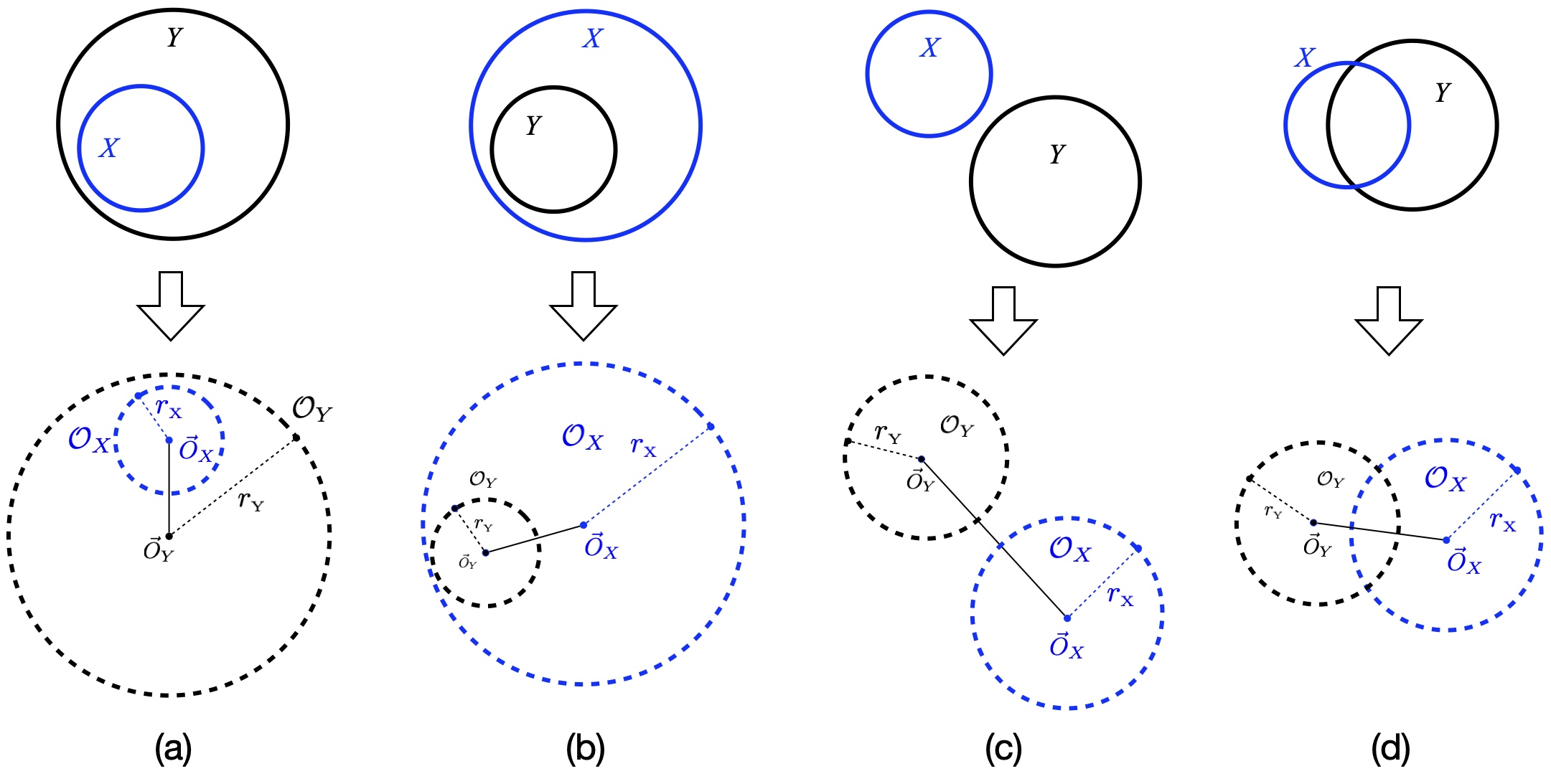}
\caption{Set relations: (a) $X\subseteq Y$, (b) $Y\subseteq X$, (c) $X\cap Y=\emptyset$, and (d) $X\cap Y\neq\emptyset\land X\nsubseteq Y \land Y\nsubseteq X$.
}
\label{ed}
\end{figure}

Spherical regions with non-zero radius can explicitly embed set-theoretic relations in the vector space. Given two sets $X$ and $Y$, we embed them as two spheres $\mathcal{O}_X$ (with the centre $\vec{O}_X$ and the radius $r_X$) and $\mathcal{O}_Y$ (with the centre $\vec{O}_Y$ and the radius $r_Y$), respectively. We assume that each concept denotes a set. $X$ being a subset of $Y$ ($X\subseteq Y$) is represented as $\mathcal{O}_X$ being part of $\mathcal{O}_Y$, written as $\mathbf{P}(\mathcal{O}_X, \mathcal{O}_Y)$, defined as $\mathbf{P}(\mathcal{O}_X, \mathcal{O}_Y)\triangleq \|\vec{O}_X\vec{O}_Y\| + r_X \le r_Y$; $X$ disjoint from $Y$ ($X\cap Y=\emptyset$) is represented as $\mathcal{O}_X$ being disconnected from $\mathcal{O}_Y$, written as $\mathbf{D}(\mathcal{O}_X, \mathcal{O}_Y)$, defined as $\mathbf{D}(\mathcal{O}_X, \mathcal{O}_Y)\triangleq \|\vec{O}_X\vec{O}_Y\| > r_X + r_Y$. $\mathcal{O}_X$ partially overlapping $\mathcal{O}_Y$ is defined by $\mathbf{PO}(\mathcal{O}_X, \mathcal{O}_Y)\triangleq \neg\mathbf{P}(\mathcal{O}_X, \mathcal{O}_Y)\land\neg\mathbf{P}(\mathcal{O}_Y, \mathcal{O}_X)\land\neg\mathbf{D}(\mathcal{O}_X, \mathcal{O}_Y)$, as illustrated in Figure~\ref{ed}. 

Aristotelian syllogistic reasoning \citep{jeffrey81} involves two premises and a conclusion using three terms, for example:
\syllogism[]{ Some {\em Greeks} are {\em rich}.\\ No {\em Philosophers} are {\em rich}.}{Some
{\em Greeks} are not {\em Philosophers}.}  
The common concept {\em rich} in the premises establishes the relation between {\em Greeks} and {\em Philosophers}. The four statement forms are: (1)~{\em universal affirmative}: all $X$ are $Y$; (2)~{\em particular affirmative}: some $X$ are $Y$; (3)~{\em universal negative}: no $X$ are $Y$; (4)~{\em particular negative}: some $X$ are not $Y$. 
Four relations in Aristotelian syllogistic statements can be explicitly embedded as follows: {\em All X are Y} is embedded as $\mathbf{P}(\mathcal{O}_X, \mathcal{O}_Y)$; {\em No X are Y} is embedded as $\mathbf{D}(\mathcal{O}_X, \mathcal{O}_Y)$; {\em Some X are Y} is embedded as $\neg\mathbf{D}(\mathcal{O}_X, \mathcal{O}_Y)$; and {\em Some X are not Y} is embedded as $\neg\mathbf{P}(\mathcal{O}_X, \mathcal{O}_Y)$~\cite{djl2025}.   

\section{Symbolic-Level Neural Reasoning through Map-guided Model Construction without Training}
\label{neural_diagrammatic_reasoning}

The central notion of logical reasoning, from the origin of logic research till now, is the notion of ``following from”, or more formally, ``logical consequence from the premises”: what can we know from the premises?~\citep{historyLogic17}. 
A logical argument can be {\em satisfiable, unsatisfiable, valid}, or {\em invalid}. Being {\em satisfiable} means there is a case in which both the premises and the conclusion are true. Being {\em valid} means the conclusion is true in every case its premises are true \citep{jeffrey81}. An argument is {\em valid} if and only if its premises together with the negation of its conclusion are {\em unsatisfiable}. 
These notations can be interpreted in terms of sphere configurations as follows: A syllogistic argument is {\em satisfiable} if and only if we can construct a sphere configuration that satisfies both the premises and the conclusion; otherwise, it is {\em unsatisfiable}. A syllogistic argument is {\em invalid} if and only if we can construct a sphere configuration that satisfies the premises and the negation of the conclusion; otherwise, it is {\em valid}. In continuous space, we can realise the construction by transforming the current sphere configuration towards the target. 
\begin{figure}[!t] 
\centering
\includegraphics[width=0.46\textwidth]{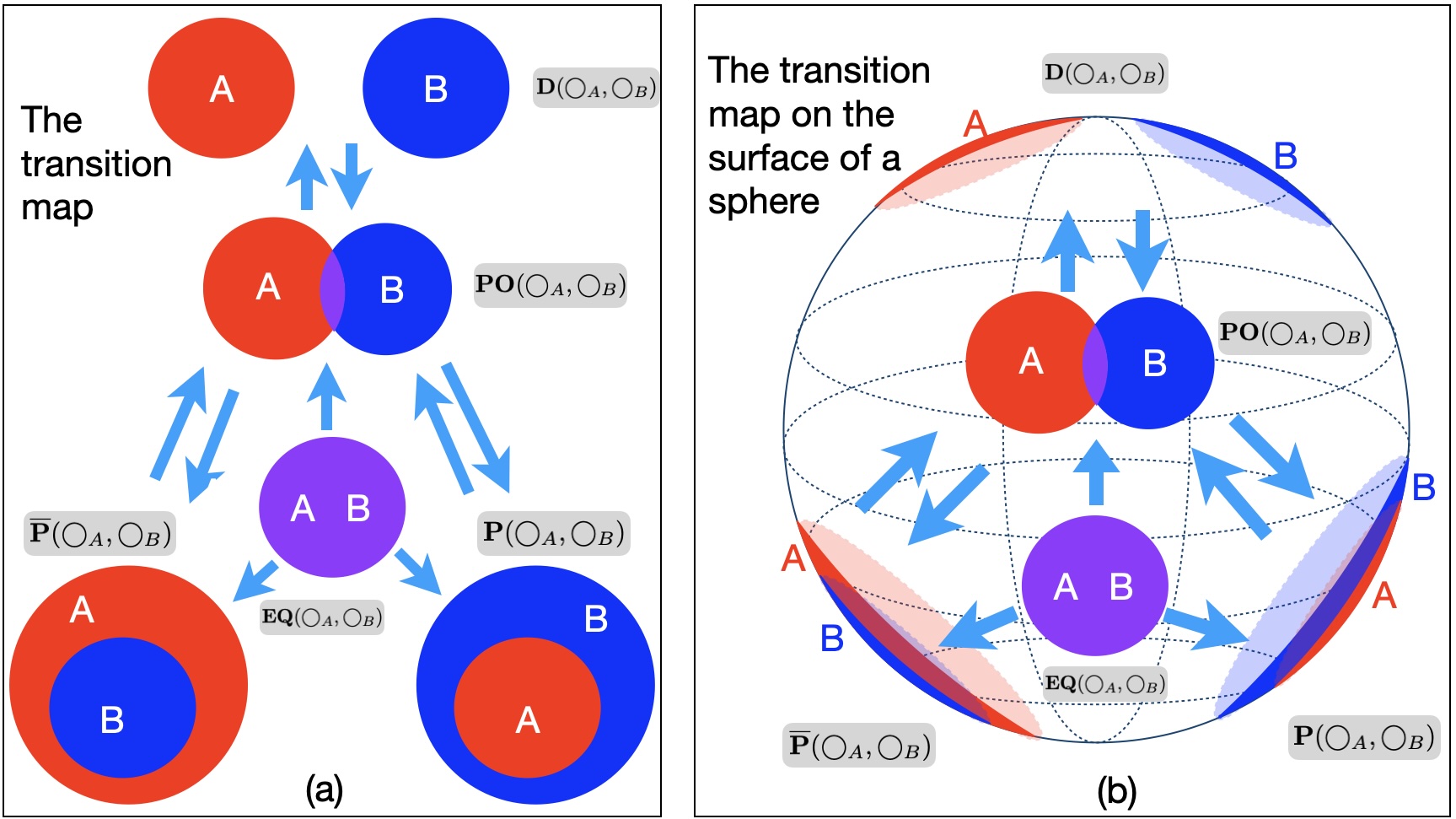}
\caption{The neighbourhood transition map in (a)~$n$-di\-mensional space and (b) on the surface of a sphere.}
\label{fig:map}
\end{figure}

During the transformation of a sphere from {\em disconnecting with} another sphere to {\em being part of} it, the two spheres shall first {\em partially overlap with} each other. The {\em being-partially-overlapped} relation is a neighbour of the {\em disconnecting with} relation. Integrating all neighbourhood transitions together, we have a transition map for all possible neighbourhood transitions, as illustrated in Figure~\ref{fig:map}(a). Rather than being trained on data, SphNNs are equipped with this transition map. To continue the example above, SphNNs first translate the premises and the conclusion into the sphere configuration: \syllogism[]{$\neg\mathbf{D}(\mathcal{O}_{\mbox{Greeks}},\mathcal{O}_{\mbox{rich}})$.\\ $\mathbf{D}(\mathcal{O}_{\mbox{Philosophers}},\mathcal{O}_{\mbox{rich}})$.}{$\neg\mathbf{P}(\mathcal{O}_{\mbox{Greeks}},\mathcal{O}_{\mbox{Philosophers}})$.}
To test whether the argument is {\em valid}, 
SphNN tries to construct a counterexample: a configuration of $\mathcal{O}_{\mbox{Greeks}}$, $\mathcal{O}_{\mbox{Philosophers}}$, and $\mathcal{O}_{\mbox{rich}}$, satisfying $\neg\mathbf{D}(\mathcal{O}_{\mbox{Greeks}},\mathcal{O}_{\mbox{rich}})$, $\mathbf{D}(\mathcal{O}_{\mbox{Philosophers}},\mathcal{O}_{\mbox{rich}})$, and $\mathbf{P}(\mathcal{O}_{\mbox{Greeks}},\mathcal{O}_{\mbox{Philosophers}})$. Then, SphNN begins its construction process: it creates three coinciding spheres, fixes $\mathcal{O}_{\mbox{Philosophers}}$, moves $\mathcal{O}_{\mbox{rich}}$ away from $\mathcal{O}_{\mbox{Philosophers}}$ till the two spheres disconnect from each other, and moves $\mathcal{O}_{\mbox{Greeks}}$ inside $\mathcal{O}_{\mbox{Philosophers}}$. This causes $\mathcal{O}_{\mbox{Greeks}}$ to disconnect from $\mathcal{O}_{\mbox{rich}}$, which contradicts the first condition and the global loss is greater than 0. The complete construction process is a well-designed, gradual descent procedure that guarantees that {\em for any satisfiable set of syllogistic constraints, SphNN constructs a satisfying sphere configuration in one epoch}~\cite{djl2024sphere,djl2025}. By this guarantee, if SphNN completes one epoch without constructing a counterexample, it concludes that the argument is {\em valid}.

\section{Summary and Outlook}

The vector representations used in conventional neural networks can be understood as a degenerate case of SphNN sphere representations in which every radius is zero. This restriction collapses regions into points, eliminating the geometric relations required to define the transition map. Consequently, this representation lacks the region-based transition mechanism used by SphNNs for syllogistic inference, while conventional neural networks typically learn to approximate such reasoning from training data. From this perspective, the McCulloch–Pitts model imposes a fundamental oversimplification: setting all radii to zero. SphNNs restore this missing structure, thereby enabling logical transitions to be encoded directly in the model rather than learned indirectly from examples.
Geometrically, a sphere in a given universe can be understood as the set of all points whose distances to a fixed point (the centre of the sphere) do not exceed a constant value (the radius). If the universe is a circle, spheres become arcs. The two first-order expressions $\forall x\ force\_in\_space(x)$ and $\forall x\ curvation\_in\_spacetime(x)$ define two sets: the set of all forces in space, and the set of all curvation in spacetime. We can embed them as two spheres on a circle, so their complements are arcs as well. To reason with disjunctive syllogism, we shall define the transition map on the surface of a sphere, as shown in Figure~\ref{fig:map}(b), 
\begin{figure}[!t]
\centering
\hspace*{-2mm}\includegraphics[width=0.48\textwidth]{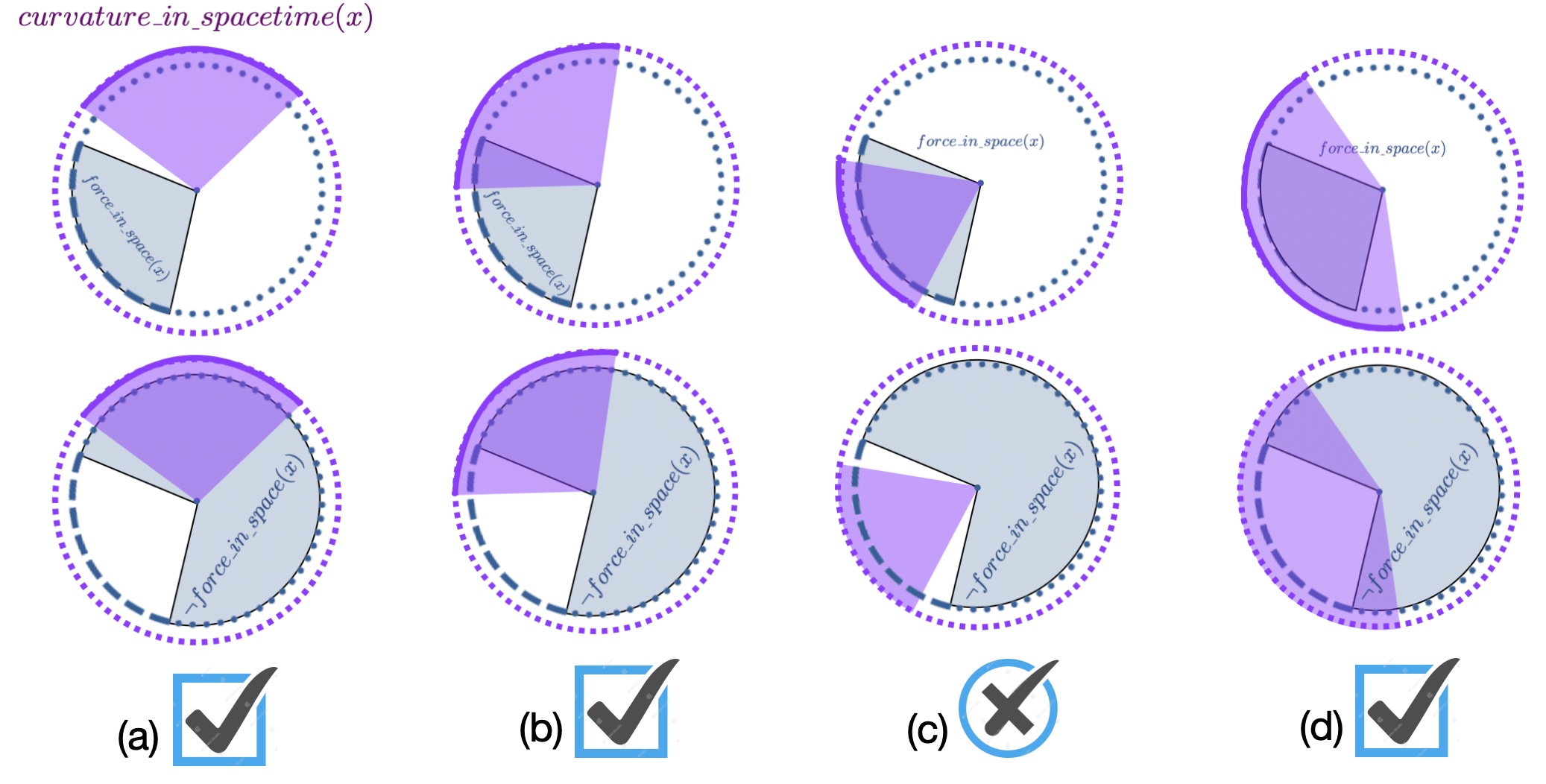}
\caption{Embedding a set as arcs allows the embedding of its complement also to be an arc.}
\label{fig:arc_sphere}
\end{figure}
so that SphNNs can update the arc configuration of sets or their complements and construct arc configurations for the additional 16 types of syllogistic-style reasoning~\cite{critical_thinking21}, including disjunctive syllogistic reasoning -- Figure~\ref{fig:arc_sphere} illustrates three correct target configurations (only Figure~\ref{fig:arc_sphere}(c) is incorrect). The zero-radius restriction might be a neglected barrier to AGI, and removing it brings AI closer to symbolic-level rational reasoning.\looseness-1

\bibliographystyle{ACM-Reference-Format}
\bibliography{XBib_NN_s, XBib}

@UNPUBLISHED{Davis81,
 author={Davis, E.},
 title={Organizing spatial knowledge},
 note={Department of Computer Science, Yale University, Research Rep. 193},
 year={1981},
}

@article{GoodwinLaird05,
author = {Goodwin, Geoffrey and Johnson-Laird, Phil},
year = {2005},
month = {05},
pages = {468-93},
title = {Reasoning About Relations.},
volume = {112},
journal = {Psychological review}, 
}

@article{RagniMarkus13,
author = {Ragni, Marco and Knauff, Markus},
year = {2013},
month = {06},
pages = {561-588},
title = {A Theory and a Computational Model of Spatial Reasoning With Preferred Mental Models},
volume = {120},
journal = {Psychological review},
}

@article{knauffetc13,
author = { Markus   Knauff  and  Leandra   Bucher  and  Antje   Krumnack  and  Jelica   Nejasmic },
title = {Spatial belief revision},
journal = {Journal of Cognitive Psychology},
volume = {25},
number = {2},
pages = {147-156},
year  = {2013},
publisher = {Routledge}, 
}

@book{Harman1986,
	author = {Gilbert Harman},
	publisher = {Cambridge, MA, USA: MIT Press},
	year = {1986},
	title = {Change in View: Principles of Reasoning}
}

@article{GAeRDENFORS90, 
 author = {Peter G\"adenfors},
 journal = {Revue Internationale de Philosophie},
 number = {172 (1)},
 pages = {24--46},
 publisher = {Revue Internationale de Philosophie},
 title = {The Dynamics of Belief  Systems: Foundations vs. Coherence Theories},
 urldate = {2023-02-10},
 volume = {44},
 year = {1990}
}

@book{Gardenfors1988,
	title = {Knowledge in Flux. Modelling the Dymanics of Epistemic States},
	year = {1988},
	author = {P. G\"{a}rdenfors},
	publisher = {MIT Press}
}

@BOOK{Knauff13,
  AUTHOR =       {Knauff, Markus},
  TITLE =        {Space to {Reason}},
  PUBLISHER =    {MIT Press},
  YEAR =         {2013}, 
}

@article{Knauff09,
author = { Markus   Knauff },
title = {A Neuro-Cognitive Theory of Deductive Relational Reasoning with Mental Models and Visual Images},
journal = {Spatial Cognition \& Computation},
volume = {9},
number = {2},
pages = {109-137},
year  = {2009},
publisher = {Taylor & Francis},
}

@ARTICLE{DongJPL,
  AUTHOR =       {Dong, T.},
  TITLE =        {A {Comment} on {RCC}: from {RCC} to {RCC}$^{++}$},
  JOURNAL =      {Journal of {Philosophical} {Logic}},
  YEAR  =        {2008},
  volume =       {37},
  number =       {4},
  pages  =      {319-352},
}

@INCOLLECTION{GoodayCohn94,
  AUTHOR =       {Gooday, J. M. and Cohn, A. G.},
  TITLE =        {Conceptual {Neighbourhoods} in {Spatial} and {Temporal} {Reasoning}},
  BOOKTITLE =    {Proceedings {ECAI-94} {Workshop} on {Spatial} and {Temporal} {Reasoning}},
  PUBLISHER =    {Rodríguez, R},
  YEAR =         {1994},

}

@TECHREPORT{McDermott81,
  AUTHOR =       {McDermott, D.},
  TITLE =        {Finding objects with given spatial properties},
  INSTITUTION =  {Department of Computer Science, Yale University},
  YEAR =         {1981},
  note =         {Research Rep. 195.},
}

@ARTICLE{McNamara86,
  AUTHOR =       {McNamara, T. P.},
  TITLE =        {Mental {Representation} of {Spatial} {Relations}},
  JOURNAL =      {Cognitive {Psychology}},
  YEAR =         {1986},
  volume =       {18},
  pages =        {87-121},
}

@ARTICLE{StevensCoupe78,
  AUTHOR =       {Stevens, A. and Coupe, P.},
  TITLE =        {Distortions in judged spatial relations},
  JOURNAL =      {Cognitive {Psychology}},
  YEAR =         {1978},
  volume =       {10},
  numbers = {4},
  pages =        {422–437},
}

@ARTICLE{Tolman48,
  AUTHOR =       {Tolman, E. C.},
  TITLE =        {Cognitive {Maps} in {Rats} and {Men}},
  JOURNAL =      {The {Psychological} {Review}},
  YEAR =         {1948},
  volume =       {55},
  number =       {4},
  pages =        {189-208},
}

@article{pittslife,
  author  = {Neil R. Smalheiser},
  title   = {Walter Pitts},
  journal = {Perspectives in Biology and Medicine},
  volume   = {43},
  number   = {2},
  pages    = {217--226},
  year     = {2000},
}

@article{rosenblatt1958,

  title={The Perceptron: A Probabilistic Model for Information Storage and Organization in the Brain},

  author={Rosenblatt, Frank},

  journal={Psychological Review},

  year={1958}

}

@article{llama3,
      title={The Llama 3 Herd of Models}, 
      author={MetaAI},
      year={2024},
      eprint={2407.21783},
      archivePrefix={arXiv},
      primaryClass={cs.AI}, 
}

@inproceedings{biomistral2024,
    title = "{B}io{M}istral: A Collection of Open-Source Pretrained Large Language Models for Medical Domains",
    author = "Labrak, Yanis  and
      Bazoge, Adrien  and
      Morin, Emmanuel  and
      Gourraud, Pierre-Antoine  and
      Rouvier, Mickael  and
      Dufour, Richard",
    editor = "Ku, Lun-Wei  and
      Martins, Andre  and
      Srikumar, Vivek",
    booktitle = "Findings of ACL 2024",
    month = aug,
    year = "2024", 
    pages = "5848--5864", 
}

@misc{gemma2024,
      title={Gemma: Open Models Based on Gemini Research and Technology}, 
      author={Team {Gemma} and {Deepmind} {Google}},
      year={2024},
      eprint={2403.08295},
      archivePrefix={arXiv},
      primaryClass={cs.CL}, 
}

@article{LeNet, 
  author = {LeCun, Yann and Bottou, L{'e}on and Bengio, Yoshua and Haffner, Patrick}, 
  journal = {Proceedings of the IEEE},
  number = 11,
  pages = {2278–2324},
  publisher = {IEEE},
  title = {Gradient-based learning applied to document recognition},
  volume = 86,
  year = 1998
}

@article{syllogism24, 
   author = {Lampinen, Andrew K and Dasgupta, Ishita and Chan, Stephanie C Y and Sheahan, Hannah R and Creswell, Antonia and Kumaran, Dharshan and McClelland, James L and Hill, Felix}, 
   title = "{Language models, like humans, show content effects on reasoning tasks}", 
   journal = {PNAS Nexus}, 
   volume = {3}, 
   number = {7}, 
   year = {2024},    
   issn = {2752-6542},  
}

@inproceedings{djl2025,
  author    = {Tiansi Dong and Mateja Jamnik and Pietro Liò},
  title     = {Neural {Reasoning} for {Sure} {Through} {Constructing} {Explainable} {Models}},
  booktitle = {AAAI},  
  year = {2025},  
}

@article{palm22023,
      title={PaLM 2 Technical Report}, 
      author={Google},
      year={2023},
      eprint={2305.10403},
      archivePrefix={arXiv},
      primaryClass={cs.CL}, 
}

@article{mistral7b,
      title={Mistral 7B}, 
      author={Albert Q. Jiang and Alexandre Sablayrolles and Arthur Mensch and Chris Bamford and Devendra Singh Chaplot and Diego de las Casas and Florian Bressand and Gianna Lengyel and Guillaume Lample and Lucile Saulnier and Lélio Renard Lavaud and Marie-Anne Lachaux and Pierre Stock and Teven Le Scao and Thibaut Lavril and Thomas Wang and Timothée Lacroix and William El Sayed},
      year={2023},
      eprint={2310.06825},
      archivePrefix={arXiv},
      primaryClass={cs.CL}, 
}

@article{openAI23,
      title={{GPT-3.5}}, 
      author={{OpenAI}},
      year={2023},
      publisher={2Retrieved September 19th, 2023.}, 
      url={https://platform.openai.com/docs/models}, 
}

@article{mistral23,
      title={Mixtral of experts- a high quality sparse mixture-of-experts}, 
      author={{AI} {Team} {Mistral}},
      year={2023}, 
      url={https://mistral.ai/news/mixtral-of-experts}, 
}

@misc{llama22023,
      title={Llama 2: Open Foundation and Fine-Tuned Chat Models}, 
      author={Hugo Touvron, et. al.},
      year={2023},
      eprint={2307.09288},
      archivePrefix={arXiv},
      primaryClass={cs.CL}, 
}

@article{syllobio2025,
      title={{SylloBio-NLI}: Evaluating Large Language Models on Biomedical Syllogistic Reasoning}, 
      author={Magdalena Wysocka and Danilo Carvalho and Oskar Wysocki and Marco Valentino and Andre Freitas},
      year={2025},
    journal={arXiv:2410.14399},
      archivePrefix={arXiv},
      primaryClass={cs.CL},
      url={https://arxiv.org/abs/2410.14399}, 
}

@inproceedings{ImageNet_Nips12,
 author = {Krizhevsky, Alex and Sutskever, Ilya and Hinton, Geoffrey E},
 booktitle = {Advances in Neural Information Processing Systems},
 editor = {F. Pereira and C.J. Burges and L. Bottou and K.Q. Weinberger},
 pages = {},
 publisher = {Curran Associates, Inc.},
 title = {ImageNet Classification with Deep Convolutional Neural Networks}, 
 volume = {25},
 year = {2012}
}

@book{jeffrey81, 
  author = {Jeffrey, R.},     
  publisher = {New York, NY}, 
  title = {Formal logic: Its scope and limits (2nd ed.)},
  year = 1981,
}

@inproceedings{Hochreiter19rudder,
 author = {Arjona-Medina, Jose A. and Gillhofer, Michael and Widrich, Michael and Unterthiner, Thomas and Brandstetter, Johannes and Hochreiter, Sepp},
 booktitle = {NeurIPS},
 editor = {H. Wallach and H. Larochelle and A. Beygelzimer and F. d\textquotesingle Alch\'{e}-Buc and E. Fox and R. Garnett}, 
 publisher = {Curran Associates, Inc.},
 title = {RUDDER: Return Decomposition for Delayed Rewards}, 
 volume = {32},
 year = {2019}
}

@inproceedings{creswell2022selectioninference,
      title={Selection-Inference: Exploiting Large Language Models for Interpretable Logical Reasoning}, 
      author={Antonia Creswell and Murray Shanahan and Irina Higgins},
      booktitle={ICLR},
    year={2023},
}

@inproceedings{critical_thinking21,
    title = "Critical Thinking for Language Models",
    author = "Betz, Gregor  and
      Voigt, Christian  and
      Richardson, Kyle",
    booktitle = "Proceedings of the 14th International Conference on Computational Semantics (IWCS)",
    month = jun,
    year = "2021",
    address = "Groningen, The Netherlands (online)",
    publisher = "Association for Computational Linguistics", 
    pages = "63--75",
}

@inproceedings{
zelikman2022star,
title={{ST}a{R}: Bootstrapping Reasoning With Reasoning},
author={Eric Zelikman and Yuhuai Wu and Jesse Mu and Noah Goodman},
booktitle={NeurIPS},
editor={Alice H. Oh and Alekh Agarwal and Danielle Belgrave and Kyunghyun Cho},
year={2022}, 
}

@article{Ballard86, 
title={Cortical connections and parallel processing: Structure and function}, volume={9}, 
number={1}, 
journal={Behavioral and Brain Sciences}, 
publisher={Cambridge University Press}, 
author={Ballard, Dana H.}, year={1986}, 
pages={67–90}}

@article{chatgpt_nature2023, 
 author = {Biever, C.},
 journal = {Nature}, 
 pages = {686-689}, 
 title = {The easy intelligence tests that {AI} chatbots fails},
 volume = {619},
 year = {2023}
}

@article{McCulloch1943,
 author = {McCulloch, Warren S. and Pitts, Walter},
 title = {A Logical Calculus of the Ideas Immanent in Nervous Activity},
 journal = {Bulletin of Mathematical Biophysics}, 
  year = {1943}, 
  volume = {5}, 
 pages = {115-133}, 
}

@book{tversky19, 
  author = {Tversky, Barbara},     
  publisher = {Basic Books}, 
  title = {Mind in Motion},
  year = 2019,
  address = {New York, USA}
}

@book{minsky88, 
  author = {Minsky, Marvin and Papert, Seymour},     
  publisher = {MIT Press}, 
  title = {Perceptrons},
  year = 1988
}

@article{Savelli08,
  author    = {F. Savelli  and D. Yoganarasimha  and JJ. Knierim },
  title     = {Influence of boundary removal on the spatial representations of the medial entorhinal cortex},
  journal   = {Hippocampus},
  volume    = {18},
  number = {12},
  pages    = {1270-1282}, 
  year      = {2008},  
}

@article{Solstad08,
    author = {Trygve Solstad and Charlotte N. and BoccaraEmilio Kropff and May-Britt Moserand and Edvard I. Moser}, 
  publisher = {American Association for the Advancement of Science},
  title = {Representation of Geometric Borders in the Entorhinal Cortex},
  journal = {Science},
  pages = {1865-1868},
  volume = 322,
  year = 2008,
}

@article{Lever09,
    author = {Lever, C. and Burton, S. and Jeewajee, A. and O'Keefe, J. and Burgess, N.},  
  title = {Boundary vector cells in the subiculum of the hippocampal formation},
  journal = {The Journal of neuroscience},
  pages = {9771–9777},
  volume =  29,
  number = 31,
  year = 2009,
}

@inproceedings{Vaswani17,
 author = {Vaswani, Ashish and Shazeer, Noam and Parmar, Niki and Uszkoreit, Jakob and Jones, Llion and Gomez, Aidan N and Kaiser, \L ukasz and Polosukhin, Illia},
 booktitle = {Advances in Neural Information Processing Systems},
 editor = {I. Guyon and U. Von Luxburg and S. Bengio and H. Wallach and R. Fergus and S. Vishwanathan and R. Garnett},
 pages = {},
 publisher = {Curran Associates, Inc.},
 title = {Attention is All you Need}, 
 volume = {30},
 year = {2017}
}

@article{syladvertising14,
author = {Jaganathan, Paramaswari and Mayr, Siti and Nagaratnam, Florence},
year = {2014},
month = {02},
pages = {151-169.},
title = {Rhetorical Syllogism in the English and the German Language of Automobile Advertising},
journal = {Gema Online Journal of Language Studies 151},
}

@article{education15,
title = {How Do Primary School Pupils Think? Syllogistic Reasoning in Primary School Children},
journal = {Procedia - Social and Behavioral Sciences},
volume = {203},
pages = {57-62},
year = {2015},
note = {International Conference EPC-TKS 2015},
issn = {1877-0428},  
author = {Denisa Ardelean},
}

@inproceedings{LJA23,
    title = "Syllogistic Reasoning for Legal Judgment Analysis",
    author = "Deng, Wentao  and
      Pei, Jiahuan  and
      Kong, Keyi  and
      Chen, Zhe  and
      Wei, Furu  and
      Li, Yujun  and
      Ren, Zhaochun  and
      Chen, Zhumin  and
      Ren, Pengjie", 
    booktitle = "EMNLP",
    month = dec,
    year = "2023",
    address = "Singapore", 
}

@article{law23,
author = {Constant, Axel},
year = {2023},
month = {04},
pages = {1-22},
title = {A Bayesian model of legal syllogistic reasoning},
volume = {32},
journal = {Artificial Intelligence and Law},
doi = {10.1007/s10506-023-09357-8}
}

@article{alphaGo2020, 
  author = {Julian Schrittwieser and Ioannis Antonoglou and Thomas Hubert and Karen Simonyan and  Laurent Sifre and Simon Schmitt and Arthur Guez and Edward Lockhart and Demis Hassabis and  Thore Graepel and Timothy Lillicrap  and  David Silver},     journal = {Nature}, 
    pages = {604–-609},
  publisher = {Macmillan Publishers Limited, part of Springer Nature. All rights reserved.}, 
  title = {Mastering {Atari}, {Go}, chess and shogi by planning with a learned model},   
volume = 588,  
  year = 2020
}

@incollection{Rumelhart86xor,
 author = {Rumelhart, D. E. and Hinton, G. E. and Williams, R. J.},
 title  = {Learning Internal Representations by Error Propagation},
 booktitle = {Parallel Distributed Processing: Explorations in the Microstructure of Cognition, Vol. 1},
 editor = {Rumelhart, David E. and McClelland, James L. and PDP Research Group, CORPORATE},
 year = {1986}, 
 pages = {318--362},
 numpages = {45}, 
 publisher = {MIT Press}, 
}

@article{LeCunNature15,
  author    = {Yann LeCun and
               Yoshua Bengio and
               Geoffrey E. Hinton},
  title     = {Deep learning},
  journal   = {Nature},
  volume    = {521},
  number    = {7553}, 
  year      = {2015},  
}

@article {Bellmundeaat6766,
	author = {Bellmund, Jacob L. S. and G{\"a}rdenfors, Peter and Moser, Edvard I. and Doeller, Christian F.},
	title = {Navigating cognition: Spatial codes for human thinking},
	volume = {362},
	number = {6415},
	elocation-id = {eaat6766},
	year = {2018},
	doi = {10.1126/science.aat6766},
	publisher = {American Association for the Advancement of Science}, 
	journal = {Science}
}

@book{Anderson95,
author = {Anderson, James A.},
address = {Cambridge, Mass},
booktitle = {An introduction to neural networks}, 
publisher = {MIT Press},
title = {An introduction to neural networks},
year = {1995 },
}

@book{LairdByrne91,
 author = {Johnson-Laird, P. N. and Byrne, R. M. J. },
 title = {Deduction},   
 year = {1991}, 
 publisher = {Lawrence Erlbaum Associates, Inc.} 
}

@book{enigma2018, 
  author = {Mercier, Hugo and Sperber, Dan},
  publisher = {Penguin}, 
  title = {The Enigma of Reason},
  year = 2018
}

@article{
Mitchell_science.adj5957,
author = {Mitchell, Melanie},
title = {How do we know how smart {AI} systems are?},
journal = {Science},
volume = {381},
number = {6654},
pages = {adj5957},
year = {2023},
}

@incollection{Khemlani2021,
 author = {Khemlani, S.},
booktitle = {Handbook of Rationality},
title = {Psychological theories of syllogistic reasoning},
 editor = {M. Knauff and W. Spohn},
 year = {2021}, 
 publisher = {MIT Press},
 address = {Cambridge, MA, USA},
}

@article{djl2024sphere,
      title={Sphere {Neural}-{Networks} for {Rational} {Reasoning}}, 
      author={Tiansi Dong and Mateja Jamnik and Pietro Liò},
      year={2024}, 
      journal={arXiv preprint arXiv:2403.15297},
      primaryClass={cs.AI}
}

@book{keefe78, 
  author = {O'Keefe, J. and Nadel, L.},     
  publisher = {Clarendon Press}, 
  title = {Expert systemsThe Hippocampus as a Cognitive Map},
  year = 1978,
}

@Article{knauf03,
  author =   "M. Knauff and  T. Fangmeier and  C. C. Ruff and P. N. Johnson-Laird",
  year =         "2003",
  title =        "Reasoning, models, and images: behavioral measures and cortical activity",
  journal =      "Journal of Cognitive Neuroscience",
  volume =       "15",
  number =       "4", 
  pages =        "559--573",
}

@book{historyLogic17,
  title={The History of Philosophical and Formal Logic: From Aristotle to Tarski},
  author={Malpass, Alex and Marfori, Marianna Antonutti},
  year={2017},
  publisher={Bloomsbury Academic}, 
}

@misc{djl2026b,
      title={Data-driven Machine Learning Cannot Reach Symbolic-level Logical Reasoning -- The Limit of the Scaling Law}, 
      author={Tiansi Dong and Mateja Jamnik and Pietro Liò},
      year={2026},
      eprint={2606.26454},
      archivePrefix={arXiv},
      primaryClass={cs.AI},
}

@book{carnap1937,
	address = {London},
	title = {The {Logical} {Syntax} of {Language}},
	publisher = {Routledge},
	author = {Carnap, Rudolf},
	year = {1937}, 
}

@book{hilbert38,
address = {Berlin},
author = {Hilbert, David and Ackermann, Wilhelm},
edition = {2., verbesserte Aufl.}, 
publisher = {Springer}, 
title = {Grundzüge der theoretischen Logik},
year = {1938},
}

@book{russell25, 
  author = {Whitehead, Alfred North and Russell, Bertrand}, 
  publisher = {Cambridge University Press}, 
  title = {Principia Mathematica},
  year = {1925--1927}
}

@InProceedings{Eisape2024,
      title={A Systematic Comparison of Syllogistic Reasoning in Humans and Language Models}, 
      author={Tiwalayo Eisape and MH Tessler and Ishita Dasgupta and Fei Sha and Sjoerd van Steenkiste and Tal Linzen},
      year={2024}, 
  booktitle={NAACL}, 
}

@article{AlphaFold3,
  author    = {Abramson, J. and Adler, J. and Dunger, J. and et al.},
  title     = {Accurate structure prediction of biomolecular interactions with AlphaFold 3},
  journal   = {Nature}, 
  volume = {630},
  pages = {493 -- 500},
  year      = {2024}, 
}

@article{AlphaGeo,
  author    = {Trinh, Trieu H. and Wu, Yuhuai and Le, Quoc V. and   He, He and Luong, Thang},
  title     = {Solving olympiad geometry without human demonstrations},
  journal   = {Nature}, 
  volume = {625},
  pages = {476 -- 482},
  year      = {2024}
}


\end{document}